# PCIA: A Path Construction Imitation Algorithm for Global Optimization

Mohammad-Javad Rezaei[1], Mozafar Bag-Mohammadi[2*]

[1]Computer Science Group, Islamic Azad University of Kermanshah, Kermanshah, Iran, mj.rezaei@ilam.ac.ir
[2]NLP Laboratory, Ilam University, Ilam, Iran, mozafar@ilam.ac.ir

*Abstract*— In this paper, a new metaheuristic optimization algorithm, called Path Construction Imitation Algorithm (PCIA), is proposed. PCIA is inspired by how humans construct new paths and use them. Typically, humans prefer popular transportation routes. In the event of a path closure, a new route is built by mixing the existing paths intelligently. Also, humans select different pathways on a random basis to reach unknown destinations. PCIA generates a random population to find the best route toward the destination, similar to swarm-based algorithms. Each particle represents a path toward the destination. PCIA has been tested with 53 mathematical optimization problems and 13 constrained optimization problems. The results showed that the PCIA is highly competitive compared to both popular and the latest metaheuristic algorithms.

## I. INTRODUCTION

Recently, several metaheuristic algorithms have been proposed to solve hard optimization problems [1,2]. These algorithms have interesting features such as bypassing local optima, high scalability, ease of implementation, and applicability to a wide variety of engineering problems. In general, metaheuristic algorithms can be divided into five categories (see Fig. 1) based on their source of inspiration: 1- natural evolution, 2-physical phenomena of the universe, 3-the social behavior of groups of animals, 4-biological processes and structures, and 5-the social behavior of the human community.

Evolution-based algorithms are inspired by natural selection in the evolution of species. In these algorithms, the next generation is derived from the intelligent or random combination of the best individuals in the current generation. Genetic Algorithm (GA) [3], Evolution Strategy (ES) [4], Differential Evolution (DE) [5], and Genetic Programming (GP) [6] are popular examples of evolutionary algorithms. Physical-based algorithms mimic the physical principles ruling the universe. The most famous methods of this category are simulated annealing (SA) [7], Gravitational Search Algorithm (GSA) [8], Big-Bang Big-Crunch (BBBC) [9] and Memetic Algorithm (MA) [10]. The third group is based on the social behavior of a group of animals. The most famous representative of this group is Particle Swarm Optimization (PSO) [11], which mimics the social behavior of birds' flock. Other noticeable examples of this category are Ant Colony Optimization (ACO) [12], Artificial Bee Colony (ABC) [13], and Fish-Swarm Algorithm (FSA) [14].

The biological behavior of living organisms has inspired the fourth category. It includes algorithms such as Artificial Immune System (AIS) [15], Bacteria Foraging Optimization (BFO) [16], Dendritic Cell Algorithm (DCA) [17], and Krill Herd Algorithm (KHA) [18]. Finally, some methods imitate the human behavior in solving real-life problems. For example, the Imperialist Competitive Algorithm (ICA) [19] models the colonial rivalry to seize and expand their colonies. Teaching-Learning-Based Optimization (TLBO) [20] has implemented the teacher and learners learning model. Harmony Search (HS) [21] and Tabu Search (TS) [22, 23] imitate the musicians' improvisation of the harmony and neighborhood search procedure respectively.

In this paper, we have imitated the human behavior in constructing new paths to reach various destinations. We introduced a new method, called Path Construction Imitation Algorithm (PCIA), which uses the following key ideas. First, humans usually walk along frequently used pathways. Second, if an existing route is not functional, he tries to access the destination via an alternative path by modifying some parts of the route. In addition, human naturally combines local and partial paths with new paths to reach an unknown destination.

In PCIA, each particle models the human behavior searching the solution space for the optimum path. The particle represents a path toward the destination. The initial population is generated randomly. Then, PCIA constructs a new generation using the similarities and dissimilarities between short and long routes in the current iteration. Therefore, PCIA is a hybrid method that imitates the social behavior of humans (modeled by particles' behavior) in finding the best path to the destination. Hence, it is similar to both human-inspired and swarm-based methods. It also could be categorized as an evolutionary algorithm since it mixes existing paths to construct a better route.

PCIA makes new routes by merging existing paths cleverly. For example, consider two similar paths *P1* and *P2*. Also, assume that *P1* is a short path and *P2* is a long path. Probably, the goodness of *P1* is due to its differences with *P2*. Hence, the different parts of *P1* and *P2* must be preserved and their similar parts must be modified. Now, suppose that both paths are short. Most likely, the excellent performance of both paths is due to their similar parts which should be preserved in future generations. But, their dissimilar sub-paths should be changed. If both *P1* and *P2* are long, their similar and dissimilar parts must be modified and preserved respectively.

[*] Corresponding author: Mozafar Bag-Mohammadi, mozafar@ilam.ac.ir

In the next section, PCIA is introduced. In the third section, PCIA is compared with popular optimization methods. Finally, section four concludes the paper.

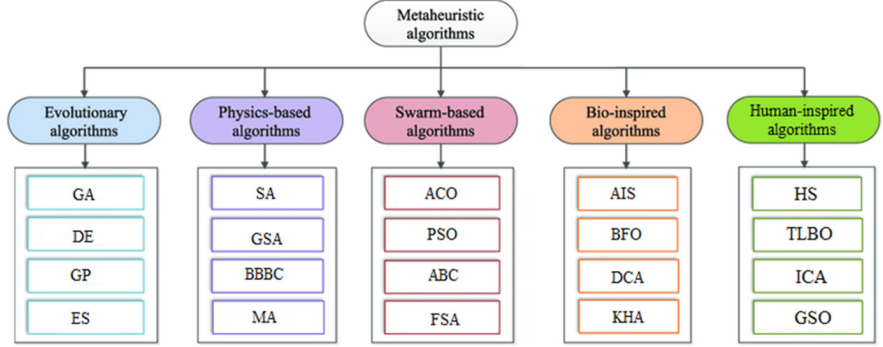

Figure 1. Classification of Metaheuristic algorithms.

## II. PCIA

PCIA imitates human behavior in exploring and constructing new paths or using existing ones. For example, humans frequently reuse popular paths. But in the case of a path closure or congestion, alternative sub-paths are activated or constructed from available routes. When constructing a new pathway to an unknown destination, the sub-paths of previous routes are examined and combined. The cost of the newly built path is an essential factor that determines its lifespan. Clearly, lengthy paths are obsoleted quickly. In contrast, a newly formed low-cost path is frequently used both individually and/or as a part of a longer path to reach another destination. Hence, its popularity is increased over time. We introduce the essential PCIA's exploitation mechanisms which are inspired by human behavior in next subsection. Then, we explain PCIA alongside its exploration mechanisms.

### A. Exploitation mechanisms

In the exploitation phase, the algorithm selects, filters, and combines current solutions to reach the promising areas of the solution space near local optima or global optimum. In PCIA, each particle represents a path toward the destination. For example, the path vector [2 -3 4 1] is a path which is composed of four sub-paths. For the route between the first and second midpoints, one should use value -3 to reach the destination. As another example, the path vectors [1.5 -2 3 1] and [2 -3 4 1] has a similar subpath (i.e. their last hop) and three dissimilar subpaths. Furthermore, we say [1.5 -2 3 1] is a better path than [2 -3 4 1] if its cost is less according to the target cost function.

PCIA classifies existing paths into two broad categories i.e. *long* and *short* path based on their cost. It then intelligently combines short and long paths to construct better routes. It selects two random paths $\vec{X}$ and $\vec{Y}$ and combines them based on their similarities and dissimilarities. Hence, the first challenge is to find a way for comparing individual elements of two paths i.e. $X[i]$ and $Y[i]$ for all $i$. The similarity criterion for two elements is defined as follows:

$$SIM_{XY}[i] = 1 - \frac{|X[i] - Y[i]|}{R[i]} \quad (1)$$

In Equation (1), $R[i]$ represents the current range of the $i^{th}$ variable. The distance between elements i.e. $X[i] - Y[i]$ is normalized by the maximum distance between any two arbitrary elements in the $i^{th}$ dimension. If the difference is zero, $SIM_{XY}[i] = 1 - \left|\frac{0}{R[i]}\right| = 1$ which indicates that the elements are completely similar. In contrast, when their difference is maximal, $SIM_{XY}[i] = 1 - \left|\frac{R[i]}{R[i]}\right| = 0$. Therefore, they are considered completely dissimilar.

Now, suppose that PCIA selected two short paths, namely $\vec{s}$ and $\vec{s'}$. In order to create an offspring, their similarities should be preserved and their dissimilarities should be reformed randomly with regard to the variables range. The similar parts are considered as the main reason behind their high performance and hence preserved. But, the different parts are modified randomly seeking for possible better routes. We use following equation for generating a new path from two short paths:

$$p_{new}[i] = \begin{cases} s[i] + r \cdot \frac{R[i]}{2} & if\ SIM_{ss'}[i] < th \\ s[i] & if\ SIM_{ss'}[i] > th \end{cases} \quad (2)$$

where $r$ is a uniform random number in [-1,1] and *th* is a predefined threshold. In fig. 2, the combination process is depicted for two short paths.

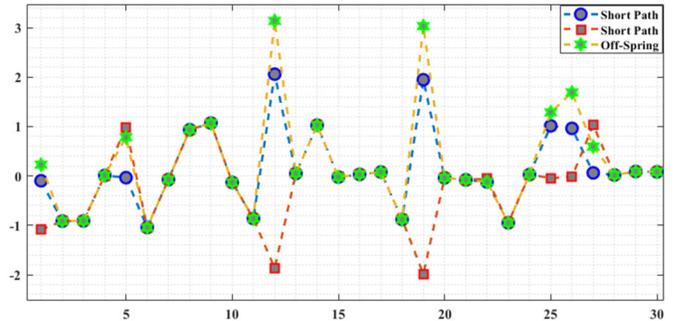

Figure 2. The combination of two short paths

Now, suppose that PCIA selected a short path $\vec{s}$ and a long path $\vec{l}$. If $SIM_{sl}[i]$ is more than the similarity threshold, $s[i]$ is modified with regard to $l[i]$ as shown in Equation (3). As this part did not help $\vec{l}$, it will not help $\vec{s}$ either. In another word, a short path learns inappropriate sub-paths from long paths and avoids them. A sample case for combination of a short and a long path is shown in Fig. 3.

$$p_{new}[i] = \begin{cases} s[i] + r \cdot \frac{R[i]}{2} & if\ SIM_{ss'}[i] > th \\ s[i] & if\ SIM_{ss'}[i] < th \end{cases} \quad (3)$$

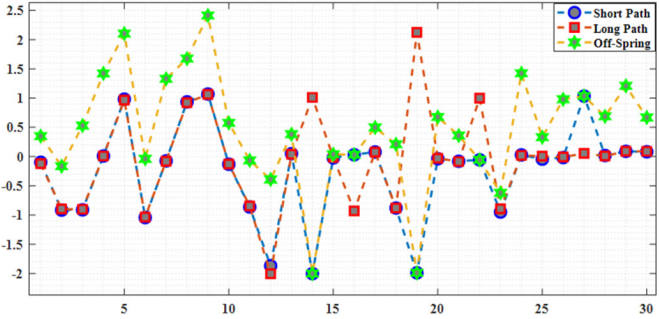

Figure 3. The combination of a short path with a long path.

For two long paths such as $\vec{l}$ and $\vec{l'}$, PCIA assumes that their similar parts are the major culprit for their subpar performances. Therefore, the similar elements are modified and the different elements are preserved using Equation (3).

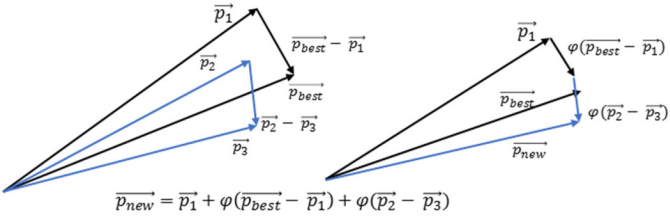

Figure 4. Illustration of assimilation strategy as used by PCIA.

The assimilation strategy used in PCIA is adopted from JADE[24] and depicted in fig. 4. It selects three random paths i.e. $\vec{p_1}$, $\vec{p_2}$, and $\vec{p_3}$. First, it rotates $\vec{p_1}$ toward $\vec{p_{best}}$ with the help of a random factor $\varphi$. Then, the resulted vector is corrected by the difference of remaining paths i.e. $\vec{p_2}$ and $\vec{p_3}$. The resulted mutant path is derived using Equation (4).

$$\vec{p_{new}} = \vec{p_1} + \varphi(\vec{p_{best}} - \vec{p_1}) + \varphi(\vec{p_2} - \vec{p_3}) \quad (4)$$

Finally, PCIA uses an exclusive smoothing strategy as shown in Equation (5). This strategy is inspired by the definition of the mathematical derivation. In the classic definition of derivation, $f'(x)$ is approximated by $\frac{f(x+\Delta x)-f(x)}{\Delta x}$. Similarly, one can use $\frac{f(x+\Delta x)-f(x)}{f'(x)}$ as an approximation for $\Delta x$. In fact, all of aforementioned strategies attempt to speed up the convergence process by sharpening the vector space. In contrast, this strategy will flatten some members of the vector space.

$$\vec{p_{new}} = \vec{p} + \frac{Cost(\vec{p}) - Cost(f'(\vec{p}))}{f'(\vec{p})} \quad (5)$$

The PCIA pseudo code is shown in "Algorithm 1". First, it generates an initial random population. Then, it iterates through the main loop until the termination criterion is met. In this loop, PCIA updates the range vector $\vec{R}$ (see Equation (1)). It uses three constant parameters i.e. $n_r$, $n_m$, and $n_s$ which represent the number of refined, mutant, and smoothed paths respectively. Then, it generates refined paths using Equation (2) and (3) in the first inner loop. In the second and third inner loops, PCIA generates mutant and smoothed paths respectively. Finally, it performs exploration mechanisms which are described in next subsection.

Algorithm 1: PCIA
1. Initialize population at first generation
2. **while** termination criterion is not met **Do**
3.    Calculate the range of population
4.    **for** i=1: $n_r$/2
5.      Select $\vec{p_1}$ and $\vec{p_2}$ randomly from population
6.      **if** both paths are short
7.        Generate a reform path using Eq. 2
8.      **else if** both paths are long
9.        Generate a reform path using Eq. 3
10.      **else**
11.        Generate a reform path using Eq. 2
12.        Generate a reform path using Eq. 3
13.    **for** i=1: $n_m$
14.      Select $\vec{p_1}$, $\vec{p_2}$, and $\vec{p_3}$ randomly from population
15.      Generate a mutant path using Eq. 4
16.    **for** i=1: $n_s$
17.      Select $\vec{p}$ randomly from population
18.      Generate a smoothing path using Eq. 5
19.    Perform exploration mechanisms
20.    Merge population and eliminate extra population

*B. Exploration mechanisms*

In the exploration phase, the population diversity must be increased in order to explore the solution space properly and prevent the algorithm from getting stuck in a local optimum. PCIA has three exploration mechanisms to construct explorative paths and avoid local optima as follow:

1- **Crossover**: Suppose that we have two paths toward destination. We can follow the first path and switch to other path somewhere in the middle of the route. Alternatively, one can start from second path and switch to the first path. This strategy is similar to a well-known technique called crossover. PCIA selects two random paths and generates new paths as follow. Here, "." is concatenation operator.

$$\begin{aligned}\vec{p_{new1}} &= \vec{p_1}(1:c).\vec{p_2}(c+1:end) \\ \vec{p_{new2}} &= \vec{p_2}(1:c).\vec{p_1}(c+1:end)\end{aligned} \quad (6)$$

Crossover strategy is only useful when two paths are completely dissimilar. It allows PCIA to avoid local optima at early iterations where the number of dissimilar paths is high. For computing the similarity of the selected paths, PCIA uses classic cosine similarity measure i.e. $\frac{\vec{p_1} \cdot \vec{p_2}}{\|\vec{p_1}\|\|\vec{p_2}\|}$.

2- **Mutation**: PCIA chooses a random path $\vec{p}$. Then, it randomly selects one of its elements and alters it as follow:

$$p_{new}[i] = p[i] + \sigma * r \quad (7)$$

Here, $r$ is a random number with normal distribution and $\sigma$ is a fraction of maximum range of $i^{th}$ dimension. Mutation strategy is useful when population members are not very similar which happens in intermediate iterations where population is converging toward optimum point.

3- **Chaos**: In order to emulate chaos, PCIA randomly selects a path and alters a small portion of it. The adjustment amount depends on the variable range of the corresponding path element.

$$p_{new}[i] = p[i] + R[i] * sign(r) \quad (8)$$

The intellectual reasons behind chaos strategy are creating diversity, avoiding boring daily routines, searching for a better path or mere curiosity. This strategy is useful in final stages where paths are converged toward the optimum solution. Clearly, mutation and chaos strategy are very similar. In mutation strategy, the adjustment value is randomly determined since the correct direction is not known yet. But in final iterations, variable range is refined and reflects the correct direction.

## C. Restart and Selection Mechanisms

The next generation is a mixture the existing members of current population $m$ and the generated paths $m'$. PCIA path selection scheme must choose $m$ paths out of $m + m'$ available paths as the members of the next generation. We have examined well-known selection mechanisms such as static tournament selection, roulette wheel selection, and rank selection and an exclusively tailored version of dynamic tournament selection. Among them, pure elitism was the best selection strategy. First, all paths are sorted according to their costs. Then, PCIA selects $m$ members from the top of the list.

PCIA has a restart mechanism which is activated when it fails to improve the target cost function effectively for 10 consecutive iterations. The improvement threshold is 0.001%. In that case, PCIA is probably stuck in a local optimum. Therefore, it stores the current best solution, deletes current population, and creates a new random population. The restart mechanism quickly bypasses local optima when other exploratory mechanisms of the PCIA fail to avoid them.

## III. SIMULATION RESULTS

We have used 53 mathematical optimization problems for evaluating the numerical efficiency of the PCIA. Problems F1 to F23 are classical benchmark functions widely used in the optimization literature [25-28]. A summary of these functions is given in Tables 1-3. For all functions, Dim, Range, and $F_{min}$ are the problem dimension, the variables range, and the global optimal respectively. The benchmark functions F1 to F23 could be divided into three categories: unimodal (Table 1), multimodal (Table 2), and fixed-dimension multimodal (Table 3). We have also selected 30 extra benchmark functions as proposed in special session on real-parameter optimization at CEC 2017 [25]. These functions, presented in Table 4, include several very complex numerical optimization problems. In Fig. 5, the typical 2D plots of the cost function are given for some of the selected benchmark functions.

**Table 1. Unimodal benchmark functions**

| Function | Dim | Range | $F_{min}$ |
|---|---|---|---|
| $F_1(x) = \sum_{i=1}^{n} x_i^2$ | 30 | [-100,100] | 0 |
| $F_2(x) = \sum_{i=1}^{n} |x_i| + \prod_{i=1}^{n} |x_i|$ | 30 | [-10,10] | 0 |
| $F_3(x) = \sum_{i=1}^{n} (\sum_{j=1}^{i} x_j)^2$ | 30 | [-100,100] | 0 |
| $F_4(x) = max_i\{|x_i|, 1 \leq i \leq n\}$ | 30 | [-100,100] | 0 |
| $F_5(x) = \sum_{i=1}^{n-1} [100(x_{i+1} - x_i^2)^2 + (x_i - 1)^2]$ | 30 | [-30,30] | 0 |
| $F_6(x) = \sum_{i=1}^{n} ([x_i + 0.5])^2$ | 30 | [-100,100] | 0 |
| $F_7(x) = \sum_{i=1}^{n} ix_i^4 + random[0,1)$ | 30 | [-1.28,1.28] | 0 |

**Table 2. Multimodal benchmark functions**

| Function | Dim | Range | $F_{min}$ |
|---|---|---|---|
| $F_8(x) = \sum_{i=1}^{n} -x_i \sin(\sqrt{|x_i|})$ | 30 | [-500,500] | -418.9829×Dim |
| $F_9(x) = \sum_{i=1}^{n} [x_i^2 - 10\cos(2\pi x_i) + 10]$ | 30 | [-5.12,5.12] | 0 |
| $F_{10}(x) = -20\exp\left(-0.2\sqrt{\frac{1}{n}\sum_{i=1}^{n} x_i^2}\right) - \exp\left(\frac{1}{n}\sum_{i=1}^{n} \cos(2\pi x_i)\right) + 20 + e$ | 30 | [-32,32] | 0 |
| $F_{11}(x) = \frac{1}{4000}\sum_{i=1}^{n} x_i^2 - \prod_{i=1}^{n} \cos\left(\frac{x_i}{\sqrt{i}}\right) + 1$ | 30 | [-600,600] | 0 |
| $F_{12}(x) = \frac{\pi}{n}\left\{10\sin(\pi y_1) + \sum_{i=1}^{n-1}(y_i - 1)^2[1 + sin^2(\pi y_{i+1})] + (y_n - 1)^2\right\} + \sum_{i=1}^{n} u(x_i, 10, 100, 4)$ $y_i = 1 + \frac{x_i}{4} \quad u(x_i, a, k, m) = \begin{cases} k(x_i - a)^2 & x_i \geq a \\ 0 & -a \leq x_i \leq a \\ k(-x_i - a)^m & x_i \leq a \end{cases}$ | 30 | [-50,50] | 0 |
| $F_{13}(x) = 0.1\left\{sin^2(3\pi x_1) + \sum_{i=1}^{n}(x_i - 1)^2[1 + sin^2(3\pi x_i + 1)] + (x_n - 1)^2[1 + sin^2(2\pi x_n)]\right\}$ $+ \sum_{i=1}^{n} u(x_i, 5, 100, 4)$ | 30 | [-50,50] | 0 |

**Table 3. Fixed-dimension multimodal benchmark functions**

| Function | Dim | Range | $F_{min}$ |
|---|---|---|---|
| $F_{14}(x) = (\frac{1}{500}\sum_{j=1}^{25}\frac{1}{j+\sum_{i=1}^{2}(x_i-a_{ij})^6})^{-1}$ | 2 | [-65,65] | 1 |
| $F_{15}(x) = \sum_{i=1}^{11}[a_i - \frac{x_1(b_i^2+b_ix_2)}{b_i^2+b_ix_3+x_4}]^2$ | 4 | [-5,5] | 0.00030 |
| $F_{16}(x) = 4x_1^2 - 2.1x_1^4 + \frac{1}{3}x_1^6 + x_1x_2 + 4x_2^2 + 4x_2^4$ | 2 | [-5,5] | -1.0316 |
| $F_{17}(x) = (x_2 - \frac{5.1}{4\pi^2}x_1^2 + \frac{5}{\pi}x_1 - 6)^2 + 10(1-\frac{1}{8\pi})\cos x_1 + 10$ | 2 | [-5,5] | 0.398 |
| $F_{18}(x) = [1+(x_1+x_2+1)^2(19-14x_1+3x_1^2-14x_2+6x_1x_2+3x_2^2)] * [30+(2x_1-3x_2)^2 * (18-32x_1+12x_1^2+48x_2-36x_1x_2+27x_2^2)]$ | 2 | [-2,2] | 3 |
| $F_{19}(x) = -\sum_{i=1}^{4}c_i\exp(-\sum_{j=1}^{3}a_{ij}(x_j-p_{ij})^2)$ | 3 | [1,3] | -3.86 |
| $F_{20}(x) = -\sum_{i=1}^{4}c_i\exp(-\sum_{j=1}^{6}a_{ij}(x_j-p_{ij})^2)$ | 6 | [0,1] | -3.32 |
| $F_{21}(x) = -\sum_{i=1}^{5}[(X-a_i)(X-a_i)^T + C_i]^{-1}$ | 4 | [0,10] | -10.1532 |
| $F_{22}(x) = -\sum_{i=1}^{7}[(X-a_i)(X-a_i)^T + C_i]^{-1}$ | 4 | [0,10] | -10.4029 |
| $F_{23}(x) = -\sum_{i=1}^{10}[(X-a_i)(X-a_i)^T + C_i]^{-1}$ | 4 | [0,10] | -10.5364 |

**Table 4. CEC'17 benchmark functions**

| | No. | Functions | $F_{min}$ |
|---|---|---|---|
| Unimodal Functions | C1 | Shifted and Rotated Bent Cigar Function | 100 |
| | C2 | Shifted and Rotated Sum of Different Power Function | 200 |
| | C3 | Shifted and Rotated Zakharov Function | 300 |
| | C4 | Shifted and Rotated Rosenbrock's Function | 400 |
| | C5 | Shifted and Rotated Rastrigin's Function | 500 |
| Simple Multimodal Functions | C6 | Shifted and Rotated Expanded Scaffer's F6 Function | 600 |
| | C7 | Shifted and Rotated Lunacek Bi_Rastrigin Function | 700 |
| | C8 | Shifted and Rotated Non-Continuous Rastrigin's Function | 800 |
| | C9 | Shifted and Rotated Levy Function | 900 |
| | C10 | Shifted and Rotated Schwefel's Function | 1000 |
| | C11 | Hybrid Function 1 (N=3) | 1100 |
| | C12 | Hybrid Function 2 (N=3) | 1200 |
| | C13 | Hybrid Function 3 (N=3) | 1300 |
| Hybrid Functions | C14 | Hybrid Function 4 (N=4) | 1400 |
| | C15 | Hybrid Function 5 (N=4) | 1500 |
| | C16 | Hybrid Function 6 (N=4) | 1600 |
| | C17 | Hybrid Function 6 (N=5) | 1700 |
| | C18 | Hybrid Function 6 (N=5) | 1800 |
| | C19 | Hybrid Function 6 (N=5) | 1900 |
| | C20 | Hybrid Function 6 (N=6) | 2000 |
| | C21 | Composition Function 1 (N=3) | 2100 |
| | C22 | Composition Function 2 (N=3) | 2200 |
| | C23 | Composition Function 3 (N=4) | 2300 |
| | C24 | Composition Function 4 (N=4) | 2400 |
| Composition Functions | C25 | Composition Function 5 (N=5) | 2500 |
| | C26 | Composition Function 6 (N=5) | 2600 |
| | C27 | Composition Function 7 (N=6) | 2700 |
| | C28 | Composition Function 8 (N=6) | 2800 |
| | C29 | Composition Function 9 (N=3) | 2900 |
| | C30 | Composition Function 10 (N=3) | 3000 |

*A. Simulation settings*

To evaluate the competency of PCIA, four representative algorithms are selected from the presented categories in Fig. 1. The selected methods are LSHADE-cnEpSin (evolutionary), PSO (swarm-based), ICA (human-inspired), and GA (evolutionary). The population size and the number of iterations were set to 120 and 1000 for all algorithms respectively. Each algorithm was run 30 times with a random initial population. Table. 5 contains the typical settings for the selected methods as reported in corresponding papers.

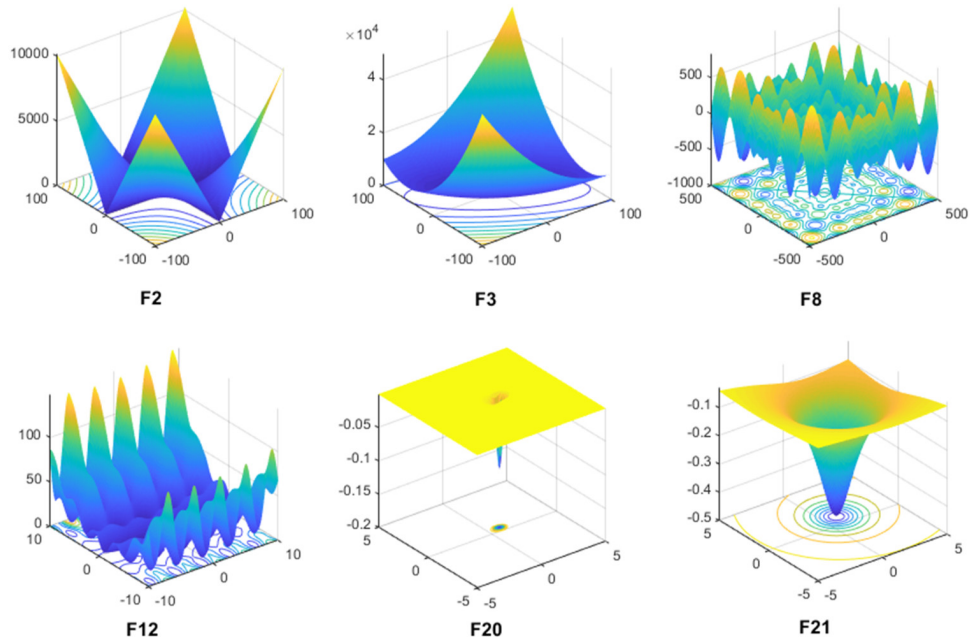

**Figure 5.** Representations of benchmark mathematical functions: Unimodal functions (F2 and F3); Multimodal functions (F8 and F12); Fixed-dimension multimodal functions (F20 and F21).

**Table 5. Details of the parameter's settings for the selected algorithms.**

| GA | PSO | ICA | LSHADE-cnEpSin |
|---|---|---|---|
| Crossover rate=0.95 | Cognition coefficient=2 | Revolution probability=0.02 | Probability best rate=0.11 |
| Mutation rate=0.05 | Personal coefficient=2 | Assimilation coefficient=2 | Arc rate=1.4 |
| Selection method=random | Inertia weight=1 | Revolution rate=0.03 | Memory size=5 |
| - | - | Number of empires=30 | Minimum population size=4 |

### B. Conventional Benchmark Functions

The results, i.e. the average cost function and the standard deviation, are reported in Table 6. The best performance for each function is shown in bold font. In 19 cases out of 23 cases, PCIA performs better than the selected algorithms. Functions F1-F7 have a single global optimum. These functions are suitable to evaluate the exploitation capability of the selected algorithms. Looking at the Rank column, it can be easily found that PCIA outperforms other algorithms for these functions. Therefore, it can be said that the exploitation capability of PCIA is superior to other methods.

In contrast to F1-F7, F8-F23 have multiple local minimums and the number of local minimums increases exponentially with the dimension of the functions. Therefore, these functions challenge the exploration capability of the selected algorithms as well as their convergence toward global optimum. The reported values for F8-F23 in Table 6 show that the exploration capability of PCIA is also very high. In fact, it outperforms other methods for 13 functions out of 15 functions.

**Table 6. Optimization results obtained for the unimodal, multimodal, and fixed-dimension multimodal benchmark functions.**

| FUN | | PCIA | GA | LSHADE-cnEpSin | PSO | ICA | Rank |
|---|---|---|---|---|---|---|---|
| F1 | avg | **8.9E-76** | 1.7E-40 | 3.79E-52 | 8.2E-23 | 7E-16 | 1 |
| | std | **2.1E-75** | 1.4E-40 | 8.56E-52 | 2.6E-22 | 2E-15 | |
| F2 | avg | **2.7E-26** | 3.8E-25 | 6.18E-24 | 7E-09 | 2E-12 | 1 |
| | std | **2.5E-26** | 1.5E-25 | 1.82E-23 | 4E-08 | 9E-12 | |
| F3 | avg | 5.1E-05 | 76.96 | **4.47E-11** | 127.19 | 0.95 | 2 |
| | std | 1.6E-04 | 129.06 | **9.66E-11** | 44.74 | 0.59 | |
| F4 | avg | **7.3E-54** | 0.51 | 9.72E-09 | 2.37 | 1.06 | 1 |
| | std | **1.5E-53** | 0.24 | 2.43E-08 | 0.6 | 0.56 | |
| F5 | avg | **0.38** | 38.86 | 13.57 | 39.09 | 32.17 | 1 |
| | std | **0.32** | 25.07 | 10.38 | 32.64 | 27.36 | |
| F6 | avg | **0** | **0** | 7.19E-34 | 2.6E-21 | 3E-16 | 1 |
| | std | **0** | **0** | 1.55E-33 | 1.2E-20 | 9E-16 | |
| F7 | avg | 0.0032 | 0.0032 | **0.0020** | 0.0102 | 0.0227 | 2 |
| | std | 0.0010 | 0.0008 | **0.0007** | 0.0033 | 0.0076 | |
| F8 | avg | **-12163** | -10917 | -12129.28 | -6104.2 | 509.81 | 1 |
| | std | **432.60** | 329.65 | 177.65 | 730.76 | 509.81 | |
| F9 | avg | 9.81 | 0.72 | **0.42** | 27.19 | 93.42 | 3 |
| | std | 3.32 | 1.01 | **0.41** | 8.96 | 16.59 | |
| F10 | avg | 5.6E-15 | **5.0E-15** | 6.10E-15 | 1.0E-07 | 4E-09 | 2 |
| | std | 1.7E-15 | **1.3E-15** | 1.80E-15 | 5.6E-07 | 1E-08 | |
| F11 | avg | **0** | 0.00074 | 5.75E-04 | 0.02 | 0.01 | 1 |
| | std | **0** | 0.00226 | 2.21E-03 | 0.03 | 0.01 | |
| F12 | avg | **1.5E-32** | **1.5E-32** | **1.5E-32** | 6.5E-23 | 5E-09 | 1 |
| | std | **5.5E-48** | 4.0E-35 | 5.89E-35 | 3.3E-22 | 1E-08 | |
| F13 | avg | 1.3E-32 | 0.0003 | 1.41E-32 | 2.4E-16 | 1E-10 | 1 |
| | std | **5.5E-48** | 0.002 | 7.05E-34 | 1.0E-15 | 3E-10 | |
| F14 | avg | **0.998** | **0.998** | 5.19 | 1.29 | **0.998** | 1 |
| | std | **2E-14** | **0** | 5.37 | 0.93 | 1E-16 | |
| F15 | avg | **0.0003** | 0.0007 | 0.0010 | 0.001 | **0.0007** | 1 |
| | std | **1.1E-12** | 4.8E-05 | 3.02E-03 | 0.0036 | 0.0001 | |
| F16 | avg | -1.0316 | -1.0316 | -1.0316 | -1.0316 | -1.0316 | 1 |
| | std | 2E-13 | 6E-16 | 6E-16 | 6E-16 | 6E-16 | |
| F17 | avg | 0.3978 | 0.3978 | 0.3978 | 0.3978 | 0.3978 | 1 |
| | std | 2.0E-13 | 0 | 0 | 0 | 0 | |
| F18 | avg | **3** | **3** | 3.9000 | **3** | **3** | 1 |
| | std | 9.7E-13 | 5.2E-16 | 4.9295 | 1.2E-15 | 1E-15 | |
| F19 | avg | -3.8628 | -3.8628 | -3.8628 | -3.8628 | -3.8628 | 1 |
| | std | 4.5E-11 | 2.7E-15 | 2.7E-15 | 2.7E-15 | 2.6E-15 | |
| F20 | avg | **-3.322** | -3.2982 | -3.278 | -3.2625 | **-3.322** | 1 |
| | std | 8.6E-09 | 0.04837 | 5.83E-02 | 0.06046 | **1.4E-15** | |
| F21 | avg | **-10.1530** | -7.9121 | -7.3891 | -6.3918 | -10.153 | 1 |
| | std | **1.6E-09** | 3.4819 | 3.1E+00 | 3.4613 | 5.8E-15 | |
| F22 | avg | **-10.402** | -10.148 | -7.9670 | -7.3745 | -10.402 | 1 |
| | std | **9.3E-10** | 1.3943 | 3.5E+00 | 3.3446 | 6.6E-16 | |
| F23 | avg | **-10.536** | **-10.536** | -9.0932 | -6.3746 | **-10.536** | 1 |
| | std | **1.4E-09** | **1.8E-15** | 2.9E+00 | 3.7364 | 1.5E-15 | |

### C. CEC'17

The CEC'17 functions have an interesting characteristic which challenges optimization algorithms by shifting the location of the global optimum and rotating the benchmark functions. The global optimum is randomly located on the boundary of the search space. In addition, the benchmark functions are rotated using $F(x)=f(R*x)$ where $R$ is a rotation matrix. These functions evaluate both exploitation and explorations capabilities simultaneously. The optimization algorithm must establish an appropriate balance between the exploration and exploitation capabilities in order to solve these optimization problems correctly and avoid frequent local optima.

The results for the CEC'17 functions are shown in Table 7. In 17 cases out of 30 cases, PCIA outperforms the selected algorithms. PCIA is the second-best algorithm in 11 cases. Clearly, classical methods such as GA, PSO, and ICA are not able to solve these complex problems efficiently. GA only outperforms others methods for function C6. ICA is the best method for two functions (i.e. C26 and C27). In remaining cases, PCIA and LSHADE-cnEpSin won 17 and 10 titles respectively. If we only consider PCIA and LSHADE-cnEpSin, PCIA outperforms LSHADE-cnEpSin in 19 cases out of 30 cases. Furthermore, we have compared these two methods by evaluating their best performance among all runs (i.e. 30) which has not reported in the table. PCIA was the superior method in 22 cases. Generally, PCIA could bypass local optima efficiently. Such ability derives from the fact that PCIA improves both exploration (through its various random path generation schemes) and exploitation (through combining short and long paths) in all iterations.

**Table 7. Comparison of optimization results obtained for CEC' 17 benchmark functions.**

| FUN | | PCIA | GA | LSHADE-cnEpSin | PSO | ICA | Rank |
|---|---|---|---|---|---|---|---|
| C1 | avg | 229.8978 | 2346.9289 | **100.0000** | 2998.0557 | 2946.4116 | 2 |
| | std | 217.2646 | 2728.4513 | **0.0000** | 2969.2977 | 3832.4145 | |
| C2 | avg | **206.4333** | 5.35E+11 | 4.01E+08 | 2.38E+13 | 3.70E+15 | 1 |
| | std | **13.8432** | 1.41E+12 | 1.87E+09 | 4.98E+13 | 8.41E+15 | |
| C3 | avg | 300.5269 | 69264.4628 | **300.3409** | 3172.3508 | 75753.5266 | 2 |
| | std | 1.9353 | 24491.5044 | **1.8673** | 1172.5239 | 22701.5772 | |
| C4 | avg | **442.9881** | 503.0640 | 461.4134 | 491.8397 | 480.9209 | 1 |
| | std | **26.3363** | 16.1047 | 19.6367 | 18.4008 | 28.0092 | |
| C5 | avg | **520.9150** | 548.9265 | 530.3125 | 652.1292 | 722.2645 | 1 |
| | std | **5.0938** | 15.2003 | 4.2579 | 27.9714 | 43.7471 | |
| C6 | avg | 600.0592 | **600.0000** | 600.0003 | 631.6744 | 640.9639 | 3 |
| | std | 0.0205 | **0.0000** | 0.0009 | 7.6604 | 6.4266 | |
| C7 | avg | **751.0841** | 773.4090 | 763.4581 | 879.3870 | 1016.7015 | 1 |
| | std | **5.5201** | 13.8791 | 4.0615 | 40.6946 | 67.2966 | |
| C8 | avg | **824.8105** | 851.7489 | 829.7727 | 915.5474 | 1017.1713 | 1 |
| | std | **4.9982** | 18.0905 | 3.0083 | 27.3250 | 31.6634 | |
| C9 | avg | **900.0014** | 900.2143 | 900.1567 | 3153.7320 | 7042.1707 | 1 |
| | std | **0.0004** | 0.3923 | 0.2406 | 852.7433 | 803.7775 | |
| C10 | avg | **3053.0578** | 4618.6548 | 3990.0780 | 4938.7731 | 4364.2905 | 1 |
| | std | **500.0122** | 623.1694 | 214.2920 | 727.9632 | 327.5258 | |
| C11 | avg | **1108.2712** | 1141.7441 | 1138.8284 | 1208.3976 | 1264.2964 | 1 |
| | std | **3.3828** | 34.7991 | 25.9472 | 33.0140 | 35.5402 | |
| C12 | avg | 28972.1300 | 724834.4792 | **5840.5739** | 569570.9526 | 398177.2936 | 2 |
| | std | 14889.3246 | 428965.9839 | **3366.8660** | 418094.9161 | 421476.7441 | |
| C13 | avg | 2138.5135 | 13048.1685 | **1546.8571** | 13363.0742 | 10476.4272 | 2 |
| | std | 1297.5393 | 7872.1968 | **168.3101** | 8977.2558 | 12870.7376 | |
| C14 | avg | **1436.3373** | 189098.4586 | 1451.7624 | 14154.5837 | 22978.9179 | 1 |
| | std | **13.0737** | 182573.7239 | 16.3803 | 19052.0397 | 20782.6862 | |
| C15 | avg | 1646.3080 | 5470.5518 | **1584.9805** | 5666.5549 | 6090.6981 | 2 |
| | std | 203.5947 | 5350.7195 | **45.0746** | 4958.3276 | 6593.9951 | |
| C16 | avg | 2123.5454 | 2559.0425 | **1838.1523** | 2748.3940 | 2455.1133 | 2 |
| | std | 151.8657 | 318.8480 | **146.3064** | 332.8830 | 203.7831 | |
| C17 | avg | **1767.4491** | 1991.8745 | 1780.0060 | 2243.2116 | 2166.7096 | 1 |
| | std | **41.1962** | 203.8652 | 25.6704 | 227.1664 | 172.7735 | |
| C18 | avg | 3731.9140 | 1092393.0445 | **1926.6247** | 341433.0531 | 202576.5950 | 2 |
| | std | 2225.8434 | 1656179.6874 | **74.2301** | 316615.9045 | 123640.7789 | |
| C19 | avg | **1924.5285** | 6961.1194 | 1951.4566 | 7769.3453 | 8635.6791 | 1 |
| | std | **21.8444** | 6247.1257 | 32.9881 | 8418.5521 | 11281.0214 | |
| C20 | avg | 2156.3212 | 2376.0618 | **2117.2165** | 2430.8712 | 2456.2281 | 2 |
| | std | 87.3075 | 214.4452 | **51.4553** | 165.0841 | 161.9677 | |
| C21 | avg | **2320.4846** | 2353.4334 | 2331.3401 | 2455.2620 | 2514.8838 | 1 |
| | std | **4.9893** | 19.4185 | 5.3933 | 39.5865 | 41.5567 | |
| C22 | avg | 2300.0169 | 2438.3014 | **2300.0000** | 4133.5188 | 3805.6550 | 2 |
| | std | 0.0046 | 757.5078 | **0.0000** | 2185.4248 | 1792.3573 | |
| C23 | avg | **2673.3936** | 2704.4370 | 2675.3360 | 3070.6617 | 2901.7972 | 1 |
| | std | **7.4782** | 13.8944 | 6.3223 | 105.3946 | 68.8575 | |
| C24 | avg | **2840.9886** | 2874.3330 | 2848.4953 | 3191.8043 | 3059.7734 | 1 |
| | std | **4.1713** | 13.5489 | 10.3502 | 126.9180 | 54.7779 | |
| C25 | avg | **2885.0028** | 2887.7942 | 2886.9334 | 2910.6680 | 2892.1398 | 1 |
| | std | **1.7211** | 1.0138 | 0.2052 | 21.4804 | 8.1439 | |
| C26 | avg | 3683.7746 | 4235.5107 | 3757.9060 | 5941.4597 | **3626.0101** | 2 |
| | std | 423.2772 | 164.9075 | 68.3431 | 1907.9178 | **1369.0482** | |
| C27 | avg | 3200.0070 | 3215.9947 | 3206.3449 | 3490.4296 | **3188.0550** | 2 |
| | std | 0.0001 | 9.5579 | 9.1205 | 125.9149 | **9.1959** | |
| C28 | avg | 3299.5200 | 3215.5972 | **3178.1855** | 3214.5474 | 3217.3727 | 5 |
| | std | 0.2793 | 14.8626 | **54.0678** | 18.2794 | 22.4229 | |
| C29 | avg | **3325.5362** | 3583.7052 | 3403.0876 | 3979.3259 | 3855.6773 | 1 |
| | std | **80.4384** | 167.9880 | 21.0772 | 243.7337 | 152.3662 | |
| C30 | avg | **3279.8471** | 7123.8756 | 5437.7400 | 7472.2027 | 12770.6085 | 1 |
| | std | **89.9829** | 1133.7189 | 329.8924 | 1487.2232 | 3710.2321 | |

*D. Analysis of Convergence Behavior*

In this subsection, the convergence behavior of PCIA is evaluated. A two-dimensional function (F17) is selected as a case study for this purpose. It has a local optimum and a global optimum. The global optimum of this function is at [-5,5] with the value -1.0316. The initial population size is 120 as usual. Fig. 6 depicts how the proposed algorithm converges toward the global optimum in steps 1, 4, 7, and 14. The short and long paths are shown in blue and green circles respectively. In step 7, nearly all short paths are around the local optimum. But, some of them are closer to the global optimum. In steps 14, all short paths are converged to the global optimum.

Fig. 7 depicts how the best path converges to the global optimum in the selected benchmark functions. In F1, PCIA starts from a corner point and quickly reaches the global optimum using large steps in the downward direction. In F8, PCIA walks along local optima until it finds the right sink. Then, it descends toward the global optimum quickly. F10 has a deep sink. PCIA finds the sink quickly and descends vertically toward it. Finally, the global optimum of F18 is in a flat area. PCIA uses a big step to reach the flat area. Then, it progresses slowly toward the global optimum.

The convergence plots of the investigated algorithms are given in Fig. 8. We have selected four representative functions from Tables 1-3. The convergence behaviors of PCIA for the selected unimodal and multimodal functions are the same. For F1 and F4, PCIA decreases the cost function dramatically while others improve the result slowly. Also, its accuracy is orders of magnitude better than other methods. In all plots, the vertical axis is logarithmic. For F6 and F11, PCIA converges very quickly and outperforms all methods. As it is evident from the PCIA restart frequency, PCIA is able to catch the global optimum 3 times while other methods are struggling to reach the optimum. This behavior could be seen in other functions with a different frequency. The restart frequency depends on the problem difficulty and the number of iterations. When the dynamic selection mechanism restarts, PCIA solves the problem again to find a better local optimum.

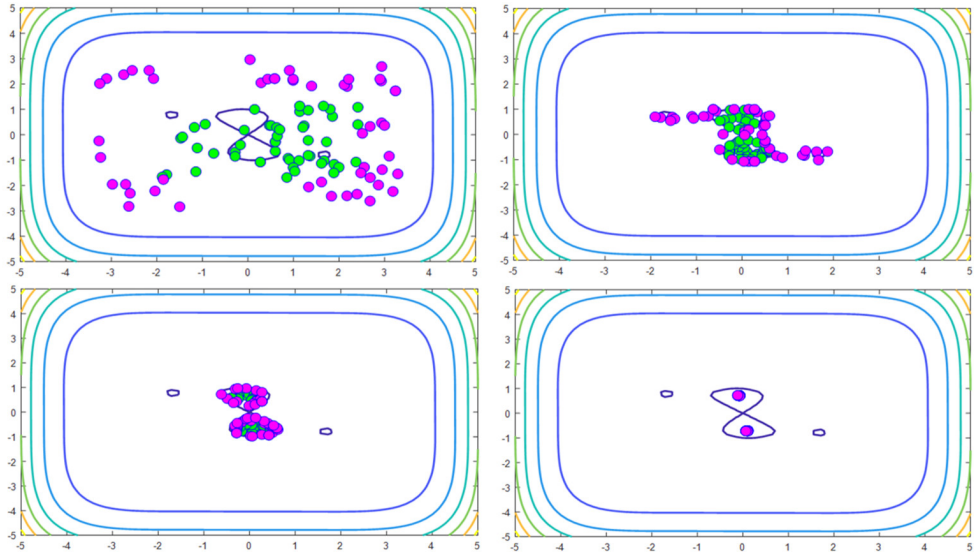

**Figure 6.** The exploitative behavior of the PCIA for F17 through various iterations (i.e. 1, 4, 7, and 14): all paths have converged to the global minimum after 14 iterations.

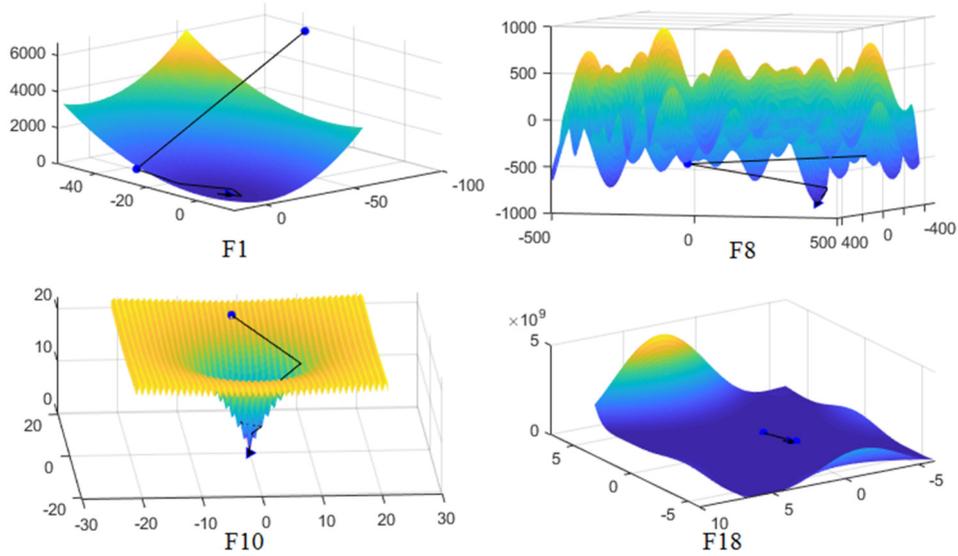

**Figure 7.** The convergence of the best particle toward the global optimum is illustrated for the selected benchmark functions.

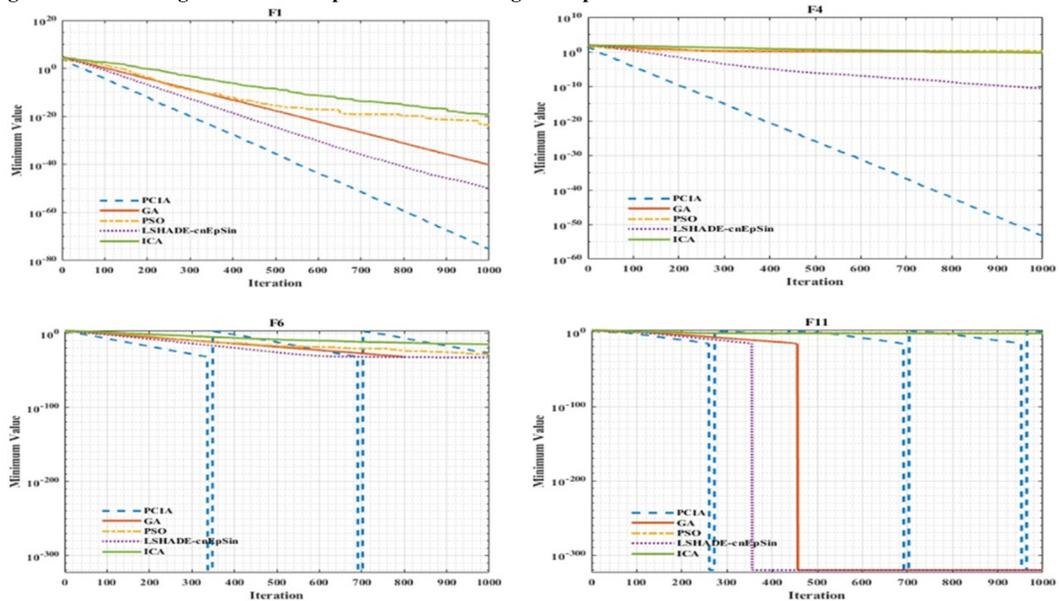

**Figure 8.** The explorative behavior of the selected algorithms for the selected mathematical benchmark functions.

### E. Constrained Functions

In this subsection, we have evaluated the effectiveness of the investigated algorithms in solving constrained optimization problems. A constrained optimization problem in the continuous space is defined as follow:

$$\min f(x)$$
$$s.t \ g_i(x) \leq 0, i = 1, \ldots, k \quad (P)$$
$$h_j(x) = 0, j = 1, \ldots, m$$
$$x \in S$$

The search space $S \subseteq R^n$ is a subset of $R^n$. It is defined as $\{x \in R^n : x_i \in [l_i, u_i], i = 1, \ldots, n\}$. Clearly, $l_i$ and $u_i$ are lower and upper bound values of $x_i$. Also, , $g_i$ and $h_j$ are real-valued functions defined on $S$. As the first step for solving this type of optimization problem, the problem constraints should be reformulated. The problem is reformulated as follow:

$$G_i(x) = (\max[0, g_i(x)])^\alpha, i = 1, \ldots, k$$
$$G_{l+j}(x) = |h_j(x)|^\alpha, j = 1, \ldots, m \quad (4)$$

where $\alpha$ is a constant integer and its value is chosen between 1 or 2. Now, the optimization problem could be solved as an unconstrained minimization problem:

$$\min_{x \in S}(f(x) + PC. \sum_{i=1}^{m+k} G_i(x)) \quad (5)$$

where PC is a *penalty coefficient* and $\sum_{i=1}^{m+k} G_i(x)$ is known as *exterior penalty function*.

For evaluation purpose, 13 multi-constraint functions are defined. The detailed formulation for these functions could be found in [26]. For each problem, PC was set high enough to prevent the constraint violation completely. It is worth noting that G2, G3, and G8 are maximization problems. These problems are converted to equivalent minimization problems and solved. The rank of PCIA is given in the last column.

The results are reported in Table 8. Maximization problems are identified by a * mark. PCIA is the most efficient algorithm in 9 cases. In the remaining 4 cases, the results are very close. For example, in G9 and G9, the differences between PCIA and the most efficient algorithm is 0.0005% and 1.79% respectively. Overall, PCIA could efficiently solve multi-constrained optimization problems.

**Table 8. Comparison of optimization results obtained for constrained optimization functions.**

| FUN | | PCIA | GA | LSHADE-cnEpSin | PSO | ICA | Rank |
|---|---|---|---|---|---|---|---|
| G1 | avg | **-15.000** | **-15.000** | **-15.000** | -13.028 | **-15.000** | 1 |
|  | std | **0.0000** | **0.0000** | **0.0000** | 1.0298 | **0.0000** |  |
| G2* | avg | **0.7962** | 0.7519 | 0.7692 | 0.3390 | 0.4536 | 1 |
|  | std | **0.0095** | 0.0341 | 0.0293 | 0.0356 | 0.0434 |  |
| G3* | avg | 0.6725 | 0.2509 | 0.4488 | **0.9845** | 0.2349 | 2 |
|  | std | 0.3161 | 0.2309 | 0.2496 | **0.0176** | 0.2085 |  |
| G4 | avg | **-30665.7** | -30523.7 | **-30665.7** | **-30665.7** | **-30665.7** | 1 |
|  | std | **0.0000** | 104.214 | **0.0000** | **0.0000** | **0.0000** |  |
| G5 | avg | **5126.88** | 5266.00 | 5346.17 | 5269.59 | 5138.64 | 1 |
|  | std | **0.3596** | 179.293 | 240.669 | 203.964 | 20.243 |  |
| G6 | avg | **-6961.81** | -6687.90 | 1653854 | -6961.79 | -6959.16 | 1 |
|  | std | **0.0000** | 175.707 | 6293318 | 0.0195 | 2.7185 |  |
| G7 | avg | **24.3075** | 25.1325 | 24.3884 | 25.6458 | 26.9612 | 1 |
|  | std | **0.0007** | 0.5036 | 0.1848 | 0.8401 | 1.4597 |  |
| G8* | avg | **0.0958** | **0.0958** | 0.0936 | 0.0914 | **0.0958** | 1 |
|  | std | **0.0000** | **0.0000** | 0.0122 | 0.0169 | **0.0000** |  |
| G9 | avg | 680.6338 | 680.8750 | **680.6304** | 680.6805 | 681.2209 | 2 |
|  | std | 0.0015 | 0.2313 | **0.0009** | 0.0293 | 0.3123 |  |
| G10 | avg | **7050.06** | 8672.00 | 7108.00 | 7505.32 | 7227.39 | 1 |
|  | std | **0.9912** | 1422.6119 | 79.2820 | 538.0880 | 151.7520 |  |
| G11 | avg | 0.7636 | 0.8199 | 0.8042 | **0.7501** | 0.7517 | 3 |
|  | std | 0.0226 | 0.0698 | 0.0979 | **0.0008** | 0.0023 |  |
| G12 | avg | **-1.0000** | **-1.0000** | **-1.0000** | **-1.0000** | **-1.0000** | 1 |
|  | std | **0.0000** | **0.0000** | **0.0000** | **0.0000** | **0.0000** |  |
| G13 | avg | 0.3986 | 0.9902 | 4.1213 | 1.1379 | **0.2610** | 2 |
|  | std | 0.2293 | 0.0430 | 11.6112 | 2.1497 | **0.1996** |  |

\* These problems are maximization problems.

## IV. CONCLUSION

A new human-inspired metaheuristic optimization algorithm has been proposed in this paper. The solution is inspired by the path construction methods in human society and the way human-being uses existing paths. The structure of the proposed method (named PCIA, path construction algorithm) stands on intelligent path refinements, various enhanced random path generations, and simple path selection. These strategies are hybridized in such a way that the PCIA explores the solution space, avoids local optima, and converges to the global optimum efficiently. The computation results for 53 well-known optimization problems (including unimodal, multimodal, hybrid, and composite functions) and 13 multi-constrained optimization problems demonstrate the superior performance of the PCIA.

## V. DATA AVAILABILITY STATEMENT

The data that support the findings of this study are available from the corresponding author, upon reasonable request.